\documentclass[a4paper, 10pt, conference]{ieeeconf}      

\IEEEoverridecommandlockouts                              

\usepackage{graphicx} 
\usepackage{amsmath}
\usepackage{amssymb}

\usepackage{bm}
\usepackage{siunitx}
\DeclareMathOperator*{\argmax}{arg\,max}

\title{\LARGE \bf
Radar-based Feature Design and Multiclass\\
Classification for Road User Recognition
}

\author{Nicolas Scheiner$^{1}$, Nils Appenrodt$^{1}$, J\"urgen Dickmann$^{1}$ and Bernhard Sick$^{2}$
\thanks{$^{1}$Daimler AG, Wilhelm-Runge-Str. 11, 89081 Ulm, Germany
        {\tt\small nicolas.scheiner@daimler.com}}
\thanks{$^{2}$Intelligent Embedded Systems, University of Kassel, Wilhelmsh\"oher Allee 73, 34121 Kassel, Germany
        {\tt\small bsick@uni-kassel.de}}%
}

\begin{document}

\maketitle
\thispagestyle{plain}  
\pagestyle{plain}  

\begin{abstract}
The classification of individual traffic participants is a complex task, especially for challenging scenarios with multiple road users or under bad weather conditions. Radar sensors provide an -- with respect to well established camera systems -- orthogonal way of measuring such scenes. In order to gain accurate classification results, 50 different features are extracted from the measurement data and tested on their performance. From these features a suitable subset is chosen and passed to random forest and long short-term memory (LSTM) classifiers to obtain class predictions for the radar input. Moreover, it is shown why data imbalance is an inherent problem in automotive radar classification when the dataset is not sufficiently large. To overcome this issue, classifier binarization is used among other techniques in order to better account for underrepresented classes. A new method to couple the resulting probabilities is proposed and compared to others with great success. Final results show substantial improvements when compared to ordinary multiclass classification.
\end{abstract}

\section{Introduction} \label{seq:intro}
Autonomous transportation is one of the driving forces in current automotive research. One major topic is adherence to safety standards. Excellent environmental perception is mandatory for this task, hence the requirements for sensors and their processing algorithms are very high. Redundancy and orthogonality of sensor systems ensure safeguarding against failure, e.g. due to hardware defects or adverse measurement conditions. Typical sensors in this context comprise cameras, radar, lidar, sonar, and others and all perception systems come with some strong points and drawbacks. In order to make use of all the different advantages of the systems, the sensor outputs are fused. Sensor fusion can occur at different stages with different effects. Roughly speaking: low-level fusion can increase accuracy as fewer information is lost. On the other hand, high-level sensor fusion improves robustness against failure of single sensors as the processing is more independent.

This work is part of the latter approach, i.e., road user classification is solely based on radar data. The big advantages of radar are the following: first, radar is the only automotive sensor that can directly measure a highly accurate radial object (Doppler) velocity at high distances. In case of a chirp sequence radar, the velocity is obtained by evaluating the phase shift between consecutive frequency chirps for a particular range. This enables single shot estimation without tracking and micro-Doppler analysis as, e.g., in \cite{schubert15} and \cite{molchanov14}. Second, the radar waves for automotive systems usually operate at the frequency bands from $24-$\SI{29}{GHz} or $76-$\SI{81}{GHz} \cite{hasch12}. To a certain extend these frequency bands allow the waves to propagate through many obstacles that would limit the propagation of optical waves, thus making them appealing for rainy or foggy situations. The weak point of automotive radar is its low angular resolution. Due to this reason, objects at higher distances have to be discriminated in range or Doppler frequency.

Classifying automotive radar data is an upcoming research topic. Many of the most popular and best performing algorithms have their origin in the field of image recognition. One reason for this is that structures such as convolutional neural networks \cite{lecun98} allow for using domain knowledge about the input data in order to implement a network structure that finds important features by itself, thus omits the task of feature engineering. With increasing computational resources, deep learning architectures such as "AlexNet" \cite{alexnet12} or "VGG" \cite{vgg14} became applicable. Unfortunately, these kinds of techniques all expect some kind of image like input.

In \cite{lomb15} and \cite{lomb17} it was shown how to use such methods for static radar data, i.e., non-moving objects such as buildings or parked cars. For dynamic data like moving road users a grid mapping procedure as in these works would require longer exposure times. Other publications such as \cite{bartsch12, heuel11}, or \cite{heuel13} use feature-based models for classification. Despite having promising results, they only consider simple dynamic situations such as the discrimination of pedestrians and static objects or vehicles. High-level autonomous driving requires understanding of complex scenes. To this end \cite{schumann17} examined the classification of several dynamic object classes using random forests and long short-term memory (LSTM) cells. Both approaches yield good performance, but the class results have a bias towards overproportionally represented classes in the training set.

The aim of this work is the overall improvement of classification results by extracting a variety of features from radar data and feeding appropriate subsets to the classifier. Therefore, features from several publications are combined and analyzed alongside some new variations. A feature selection algorithm is used to determine suitable subsets. Moreover, it is shown how the problems occurring from dataset-specific imbalance can be mitigated with different re-sampling approaches and multiclass binarization techniques such as one-vs-one (OVO) and one-vs-all (OVA). By utilizing these methods, more specialized models can be trained which achieve much better class recognition scores, mainly contributed by improvements made on minority classes. To this end a new way to couple the probabilities is proposed and compared to other existing methods.

The article is organized as follows: in Section \ref{seq:data} the data preprocessing is described and an overview over the existing dataset is given. The first part of Section \ref{seq:methods} explains how features are extracted from the data at hand and how suiting subsets are selected; the second part deals with data re-sampling and multiclass binarization techniques. Section \ref{seq:res} presents the results and Section \ref{seq:conc} concludes the topic and gives prospects for future work.

\section{Data Aquisistion} \label{seq:data}
All data for this work was acquired with four radar sensors distributed to the lateral sides and front corners of a test vehicle. The sensors' specifications are shown in Tab. \ref{tab:radar_specs}. The first row represents the carrier frequency \texttt{$f$} and the operational bands for range (distance) \texttt{$r$}, azimuth angle \texttt{$\phi$}, and radial (Doppler) velocity \texttt{$v_r$} respectively. The second row gives the resolutions in time \texttt{$\Delta t$} and for \texttt{$r$}, \texttt{$\phi$}, and \texttt{$v_r$}.
\begin{table}[htb]
\renewcommand{\arraystretch}{1.3}
	\caption{Radar sensor specifications.}
	\label{tab:radar_specs}
	\centering
	\begin{tabular}{cccc}
		\hline
 		\texttt{$f / GHz$} & \texttt{$r /m$} & \texttt{$\phi /\deg$} & \texttt{$v_r /\frac{km}{h}$}\\
		\hline\hline
		$77$ & $0.25-100$ & $\pm45$ & $-400 - +200$ \\
		\hline\hline
		\texttt{$\Delta t /ms$} & \texttt{$\Delta r /m$} & \texttt{$\Delta \phi /\deg$} & \texttt{$\Delta v_r /\frac{km}{h}$}\\
		\hline\hline
		$60$ & $0.42$ & $3.2-12.3$ & $0.43$\\
		\hline
	\end{tabular}
\end{table}

\subsection{Classification Processing Chain} 
\begin{figure}[htb]
	\centering
	\includegraphics[width=0.98\columnwidth]{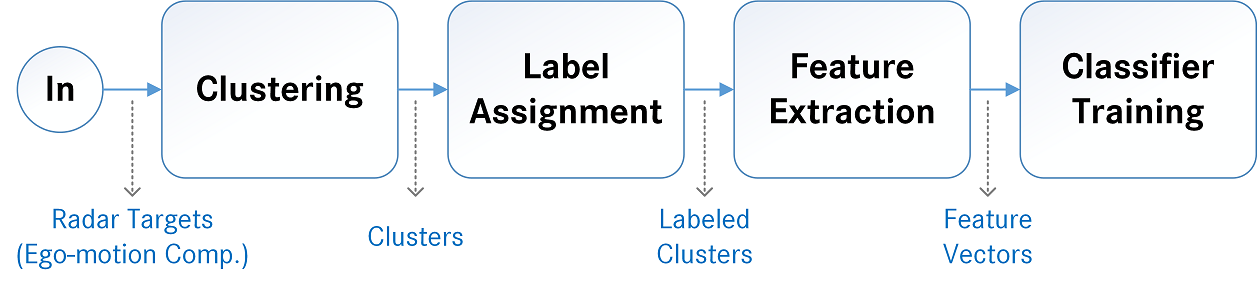}
	\caption{Simplified processing chain for classifier training. The rectangles contain the processing steps. Corresponding data levels are indicated separately.}	
	\label{fig:clf_chain}
\end{figure}

The radar sensors deliver targets which are already resolved in range, angle, and radial Doppler velocity. An internal constant false alarm rate (CFAR) \cite{richards2005} detector returns only targets which exceed an adaptive amplitude threshold. Prior to all other processing steps, all sensor data is transformed in a common coordinate system and ego-motion compensation is applied to the Doppler velocities. The data is then clustered in space, time, and Doppler using a DBSCAN \cite{ester96} algorithm. Due to the deteriorating angular resolution of the sensors towards the edges of their respective field of view, overlapping sensor regions are not treated differently than other areas. At this point, each object instance is described by a unique cluster id and passed to human experts for labeling. The labeling process also includes a manual cluster correction in order to ensure that all object targets are being considered. The uncorrected cluster and a version where $\SI{40}{\%}$ of the targets are randomly dropped are also passed to the next processing stage for data augmentation. For feature extraction, all labeled cluster sequences are first sampled in time using a window of $\SI{150}{ms}$. Then, features are extracted from each of the cluster samples. Finally, a classifier is presented with the resulting feature vectors and fits its model parameters accordingly. The simplified data pipeline is summarized in Fig. \ref{fig:clf_chain}.
During testing, the same procedure is followed, except no data augmentation is done and labels are only presented to a subsequent result evaluation, i.e., after the classifier estimated the class membership.

\subsection{Dataset}
\begin{figure}[htb]
	\centering
	\includegraphics[width=1\columnwidth]{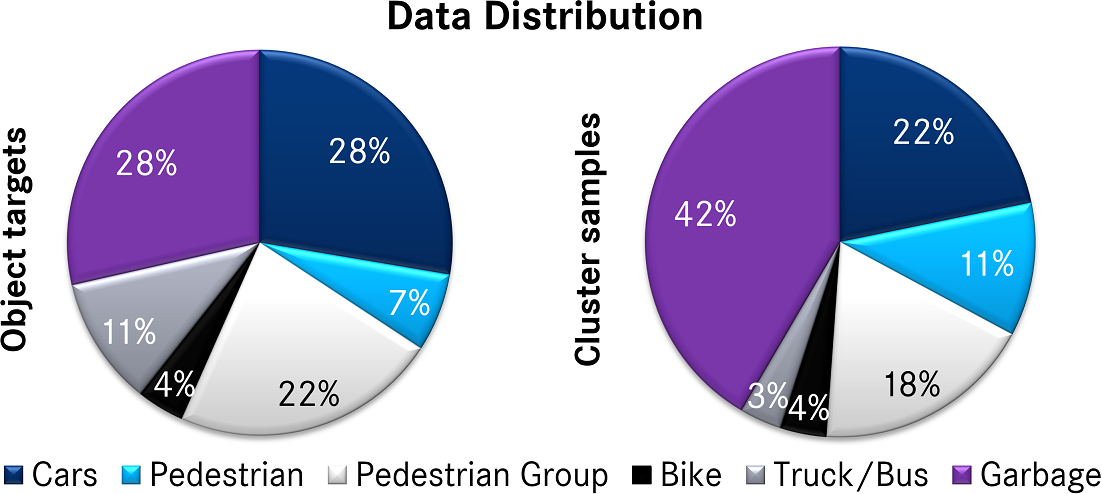}
	\caption{Dataset: distribution of object targets and cluster samples.}	
	\label{fig:dataset}
\end{figure}

The dataset examined for this work contains slightly over eleven million objects targets, which correspond to roughly one million non-augmented cluster samples of slightly over 3000 real-world objects. Object targets are all targets returned from the radar which are assigned a label in the labeling process. Hence, not every reflection seen by the radar is in the dataset, but only those which pass both the CFAR detector and the labeling process. This results in six classes: car, pedestrian, pedestrian group, bike, truck/bus, and garbage. The garbage class consists of wrongly detected and clustered measurement artifacts. An overview on the class distribution can be found in Fig. \ref{fig:dataset}.

It is clearly visible that the class distributions of both graphs in Fig. \ref{fig:dataset} are uneven. In principle, the fairest way to compare the influence of each class would be to examine the measurement time per object class, which is reflected in the cluster sample distribution. Additionally, the mean measurement duration for each class should be equalized because consecutive measurements of the same object instance usually implicate high correlations between corresponding feature vectors. Despite the described way being advantageous for treating imbalances as seen by a classifier, there is also a drawback resulting from this method because the features evaluated on small sized objects have much higher variations among different occurrences. The cause of this is the big effect of small variations in the total amount of targets, obtained for objects like pedestrians or bikes within the time frame of one cluster sample. Few targets more or less can entirely change the values in a feature vector, when the total amount of targets is very small. This effect would be negligible for a sufficiently big dataset, but for the current size, aiming at a similar class distribution on target level would also be beneficial. This makes perfect balancing challenging. Acquiring more data is possible in general, but also very expensive and time-consuming. Hence, this article is focusing only on the already existing data.

\section{Methods} \label{seq:methods}
Among several classifiers, random forests \cite{breiman01} have proven to be very reliable and fast trainable classifiers for the problem at hand. Good results are achieved using a configuration with 50 trees, Gini impurity as measure for the split quality, a maximum of features considered for splitting (the square root of the total amount of features), and no maximum on the depth of the individual trees. Experiments with more trees were carried out prior to this work, but did not result in substantial improvements when applied to the maximum feature set. A first evaluation on multiclass classification techniques used only random forests to identify beneficial configurations.
For a final comparison only a subset of promissing techniques identified by this first evaluations was used and also tested on long short-term mermory (LSTM) classifiers \cite{hochreiter97}. The LSTM model is based on \cite{schumann17}, where good results were achieved on very similar data. The model comprises a simple layer of 80 LSTM cells followed by a softmax layer to estimate the class posterior probabilities at all time steps. As input sequence, feature vectors of eight consecutive time steps are gathered in a sliding window fashion. The usage of LSTMs yields further improvements, but the substantially longer training time limits the possibility to do expensive grid searches.

\subsection{Feature Extraction}
Feature extraction is an essential step for any classification task. As stated in the introduction, great success has been made with convolutional neural networks, which learn how to extract meaningful features as part of their training routine. The random forests and LSTMs used for this work, though, profit a lot from good features. Hence, this section summarizes feature extraction methods from several publications, presents some variants, and shows how the resulting feature collection can efficiently be trimmed to a compact feature set.

\begin{figure}[htb]
	\centering
	\includegraphics[width=0.8\columnwidth]{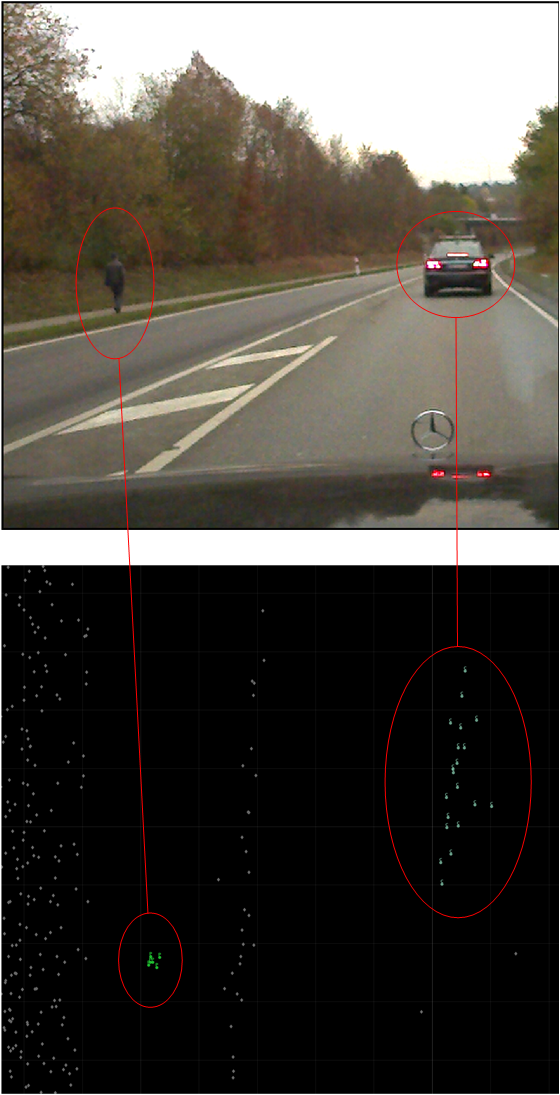}
	\caption{Excerpts from a traffic scene including a pedestrian and a car; on top: image captured by a documentation camera for verification purposes; at the bottom: spatial distribution of radar targets in cutout of the sensors' field of view accumulated over $\SI{300}{ms}$ in top-down view. The size of the time window and the radar target colors were chosen merely for visual purposes.}	
	\label{fig:scene}
\end{figure}

Without further processing, the radar sensors return values for range, angle, amplitude (considering a compensation for the free-space path loss), and radial velocity for each target. As mentioned earlier, all values are transformed to a common coordinate system before a clustering algorithm associates targets belonging to the same object instance. Features are then generated from samples of these clusters. Fig. \ref{fig:scene} depicts a typical scene of a pedestrian and a car in a rural environment. The sparsity of targets compared to urban scenes allows for a good view on the two clusters representing the two objects.

A first feature set derived from each cluster sample extracts for each of the four basic measurement values: maximum, minimum, and mean value, the standard deviation and the spread between maximum and minimum within the cluster. In accordance with the work in \cite{schumann17}, the two eigenvalues of the covariance matrix of all the x and y coordinates in each sample and its total number of targets are computed, too. The number of targets and the angle spread are also evaluated in their range-compensated version, i.e., multiplied by the mean range in order to counteract the deterioration of target recognition at high distances. To further improve the classification, the ratio of stationary to moving targets is calculated.

\begin{figure}[h]
	\centering
	\includegraphics[width=0.8\columnwidth]{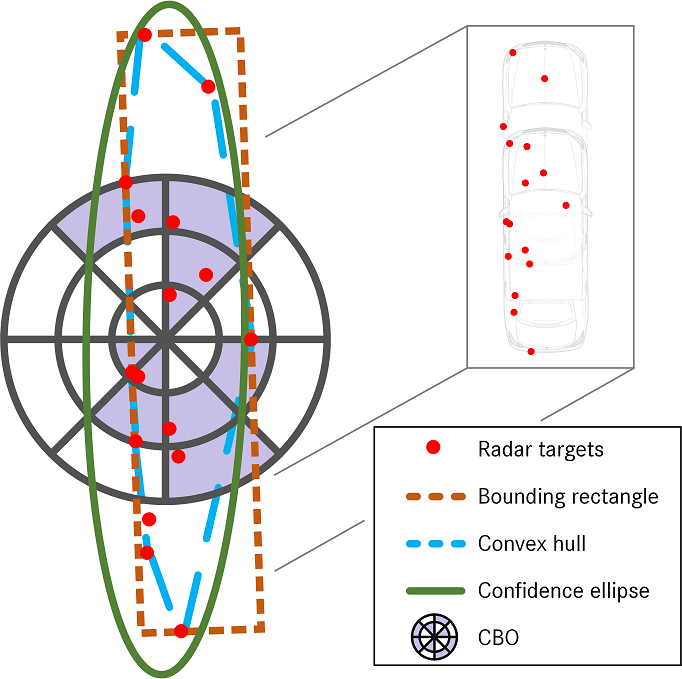}
	\caption{Geometric features calculated based on the target distribution measured from the car shown in Fig. \ref{fig:scene}. Displayed are the convex hull, the minimum bounding rectangle, the 95\% confidence ellipse and the CBO grid. In the CBO grid, the occupied sectors of each concentric bin are highlighted.}	
	\label{fig:featureset}
\end{figure}

In addition to the resulting feature set, this article includes several features concerning the shape and the distribution of clustered targets within a time frame as described in \cite{zhang16}. Experiments have been conducted for multiple features used there, in detail these include: computation of the convex hull of all targets on the x/y-plane with its area, perimeter, and target density. The isoperimetric quotient of the convex hull is used as measure of circularity, also the radius of the minimum fitting circle is calculated. Furthermore, the convex hull can easily be extended to a minimum bounding rectangle with again: area, perimeter, and target density. To better account for outliers in the cluster, the major and minor axes of a 95\% confidence ellipse are included. Another interesting feature is a keypoint descriptor, cumulative binary occupancy (CBO), introduced in \cite{zhang16} which splits the area around the center of a cluster sample (originally the median target) in concentric bins as depicted in Fig. \ref{fig:featureset}. Each of the three bins consists of eight segments of which each is examined for binary occupancy, i.e., the segment is occupied by at least one target or not. Then, for each of the three bins the number of occupied segments is tallied up to three resulting CBO values. Moreover, the average distance of all targets to the line connecting the two most distant targets of the cluster sample and the linear correlation of x and y coordinates of all targets are evaluated. The mean average distance to the cluster center is used to indicate the compactness of a cluster sample.  

Besides the geometric features, \cite{heuel13} demonstrated the usefulness of range and velocity (Doppler) profile to discriminate between cars and pedestrians. The resulting range-Doppler spread can be described as the ratio of range spread to Doppler spread. Similar to that measure, the linear correlation of range and velocity distribution can be calculated. For rotated objects the same procedure can be advantageous when spread and correlation are based on azimuth angle and Doppler. To be even less dependent on the rotation of an object, this article proposes a third variant which evaluates the target spread (and the linear correlation) in direction of minor and major axis of the confidence ellipse instead of range or angular spread.

To avoid overfitting to the so acquired feature set of 50 features in total, it is common to use only a subset of features which should contain no redundancies but still preserve important information. Feature selection is a computationally expensive task. An exhaustive search is only possible for small feature vectors as the problem is NP-hard. For this work, a backward elimination algorithm was used to find suitable feature sets. Backward elimination is a wrapper method which trains several classifiers to recursively eliminate the feature that yields the best performing classifier after leaving it out \cite{Kohavi1997}. The elimination is done in a greedy fashion, hence, once a feature is gone it will no longer be available in upcoming iterations of the algorithm. Backward elimination has a higher computational cost than the similar forward selection approach, which starts by training classifiers for all single features in a dataset and then adds one feature at a time. Despite the increased computational complexity backward selection was chosen because it is better equipped to preserve mutual information between different features. 

\subsection{Dataset Imbalance}
As shown in the previous section, the data used for the task at hand suffers from imbalances in class distribution. More specifically, it only contains few single pedestrians, bikes, and trucks. The small amount of targets is often reflected in noticeably worse classification scores for minority classes. This article aims to tackle the imbalance problem by training more specialized binary classifiers and combining their results. Moreover, classifiers are combined with a variety of different re-sampling techniques which also aim to mitigate imbalance offsets in the data. Training multiple binary classifiers rather than a single one with more than two outputs was already successfully applied to imbalanced data in \cite{fernandez13}. It allows for each classifier to specialize more on its individual task. At the same time, this provides a natural way to reduce some of the class imbalances. The challenging step is the combination of multiple classifier outputs to one meaningful result. Two of the most prominent binarization techniques are one-vs-all (OVA) and one-vs-one (OVO) classification \cite{chmielnicki2016}. OVA trains $K$ instead of one classifier where $K$ is the number of different classes in the dataset. Each classifier is fitted to separate one single class from the rest, hence the final decision for an object id  $\texttt{id}(\bm{x})$ is determined by the maximum probability $p_{i,\texttt{OVA}}(\bm{x})$ among all classifier results, where $\bm{x}$ is the feature vector of the investigated cluster:
\begin{align}
	\texttt{id}(\bm{x}) = \argmax_{i \in \{1,...,K\}} \text{\hspace{3mm}} p_{i,\texttt{OVA}}(\bm{x})\texttt{.}
	\label{eq:ova}
\end{align}
OVO on the other hand trains $K(K-1) / 2$ classifiers, one for each class pair. Several techniques have been proposed to calculate a final result from the output probabilities $p_{ij}(\bm{x})$. Friedman et al. \cite{friedman96} introduced max-voting which decides for the class with the maximum votes among classifiers. In case of a tie, the class is randomly chosen:
\begin{align}
	\nonumber \texttt{id}(\bm{x}) &= \argmax_{i \in \{1,...,K\}} \text{\hspace{1mm}} \sum_{j=1, j\neq i}^K \delta(p_{ij}(\bm{x})) \\
	&\text{\hspace{5mm} with \hspace{5mm}} \delta(p) = \epsilon(p-0.5)\texttt{.}
	\label{eq:ovo}
\end{align}
$\epsilon(p)$ denotes the Heaviside step function. As elaborated in \cite{moreira98}, $\delta(p)$ can also take many other forms in order to perform pairwise coupling (PWC) of the probabilities. According to that work, the corresponding methods are referred to as PWC-x. PWC-1 is equal to max-voting in \eqref{eq:ovo}, PWC-2 and higher use the following $\delta(p)$:
\begin{align}
	& \nonumber \text{PWC-2\hspace{5mm}} \delta(p) = p \\
	& \nonumber \text{PWC-3\hspace{5mm}} \delta(p) = (1+e^{-12(p-0.5)})^{-1} \\
	& \nonumber \text{PWC-4\hspace{5mm}} \delta(p) = p \cdot \epsilon(0.5-p) + \epsilon(p-0.5) \\
	&  \text{PWC-5\hspace{5mm}} \delta(p) = p \cdot \epsilon(p-0.5)\texttt{.}
	\label{eq:pwc}
\end{align}
As choosing randomly for a tie can occur frequently in a max-voting procedure, PWC-1 introduced a lot of uncertain decisions. This work uses an altered version which switches to PWC-2 in case of an impair.
Two very sophisticated approaches introduce correction classifiers $q_{ij}(\bm{x})$ to minimize to the influence of a classifier that was not optimized for the object type under test, e.g. when the model trained for object ids 1 vs. 3 is presented with a sample with object id 2:
\begin{align}
	\texttt{id}(\bm{x}) = \argmax_{i \in \{1,...,K\}} \text{\hspace{1mm}} \sum_{j=1, j\neq i}^K p_{ij}(\bm{x}) \cdot q_{ij}(\bm{x})\texttt{.}
	\label{eq:cc}
\end{align}
$q_{ij}(\bm{x})$ could be delivered by additional classifiers trained for samples of class $i$ and $j$ vs. the remaining ones. Similar but more efficiently, they can be the result of adding the corresponding probabilites of a seperately trained OVA classifier $q_{ij}(\bm{x}) = p_{i,\texttt{OVA}}(\bm{x}) + p_{j,\texttt{OVA}}(\bm{x})$ (PWC-OVA) \cite{moreira98}. All mentioned algorithms were tested, including an newly proposed version of PWC-OVA which applies the OVA correction not by $q_{ij}(\bm{x})$ as in \eqref{eq:cc}, but rather after a pre-selection using PWC-2. This will be referred to as PWC-OVA2:
\begin{align}
	\texttt{id}(\bm{x}) = \argmax_{i \in \{1,...,K\}} \text{\hspace{3mm}} p_{i,\texttt{OVA}}(\bm{x}) \cdot \sum_{j=1, i\neq j}^K p_{ij}(\bm{x})\texttt{.}
	\label{eq:opc}
\end{align}
Besides using different ensemble techniques for dealing with balance problems some more classical approaches were also examined, namely class weighting, undersampling, and oversampling. In contrast to binarization techniques, these methods are only applied at training time. Class weighting is used to accustom the influence of training samples inversly to their share in the total number of samples. Undersampling drops samples from the majority class(es) until an equilibrium is reached. Sample dropping can occur randomly or using fixed patterns, e.g. by removing potentialy noisy majority class samples close to the decision boundary (cf. Tomek links). Oversampling can either replicate existing data to balance the dataset or synthesize new data samples from existing ones using SMOTE, ADASYN, or variations from those algorithms. A summary of these techniques can be found in \cite{schlegel2017}. In the next section, results are compared for class weighting and all mentioned re-sampling techniques as they perform on their own or in combination with the different binarization approaches.

\section{Results} \label{seq:res}
The dataset used in this work contains strong imbalances among individual class occurrences. Since a final application for algorithms such as the ones presented here directly affects the safety of road users, all classes should have the same influence on an evaluation metric. Therefore, all classification scores are reported as macro-averaged F1-scores. The F1-score is defined as the harmonic mean of precision and recall, hence takes into account both false positives and false negatives. Macro-averaging solves the problem of imbalanced class distributions by first obtaining F1-scores for each individual class, then using the mean value of all $K$ F1-scores ($F1_\text{macro\_avg}=\frac{1}{K}\sum_{i=1}^K F1_i$) as a final measure that is equally affected by all $K$ classes.

\subsection{Feature Extraction}
Suitable feature subsets were selected following two steps: First, feature selection was used to calculate a feature ranking. Two rankings were computed using backward elimination separately for random forests and LSTMs. A normal multiclass model was used for the backward elimination algorithm. The data was split into $\SI{80}{\%}$ training and $\SI{20}{\%}$ test data and it was ensured that samples belonging to the same cluster id could never occur in both sets. For the complete ranking, backward elimination with 50 features already requires more than 1000 models to be trained, hence the 80/20 split was preferred over expensive cross-validation. In general both rankings are very similar, hence only the LSTM case is discussed here. The corresponding feature ranking can be found in Tab. \ref{tab:features}. Among the top four ranked features according to backward elimination are the length of the major axis of the 95\% confidence ellipse -- which is essentially a nonlinear mapping of the highest eigenvalue of the covariance between x and y coordinates --, the maximum radial velocity and the ratio of stationary targets within the cluster. These results align well with the results in \cite{schumann17}, where those three features (eigenvalue instead of major ellipse axis) were found to be the three most important ones. The backward elimination used in this article, however, includes the amplitude mean value and spread at rank three and six, which indicates a much higher priority than in that publication. The big difference is most likely caused by the different methods used for feature ranking. Random forest feature ranking is well equipped for single feature importance evaluation, but wrapper methods such as backward elimination usually find better feature sets as they also take into account the effect of other features on the evaluated one. Besides those basic derivations of the four measurement values (range, angle, velocity, and amplitude), the top ten features also include the perimeter of the minimum bounding rectangle and the two new variants of the spatial velocity spread along the axes of the confidence ellipse. This and consecutive features in the ranking clearly indicate the benefit gained from spatial fits and Doppler distributions.

\begin{table}[h]
\renewcommand{\arraystretch}{1.15}
	\caption{LSTM-based feature ranking (optimum subset).}
	\label{tab:features}
	\centering
	\begin{tabular}{rll}
		\hline
 		Rank & Name & Description\\
		\hline\hline
		1 & con95major & major axis length of 95\% conf. ellipse\\
		2 & vrCompMax & max. radial velocity (ego-motion comp.)\\
		3 & ampMean & mean target amplitude\\
		4 & fracStationary & percentage of stationary targets in cluster\\
		5 & nTargetsComp & target amount weighted by mean distance\\
		6 & ampSpread & difference from max. to min. amplitude\\
		7 & vrCompStd & std. dev. radial velo. (ego-motion comp.)\\
		8 & rehuPerimeter & perimeter of min. bounding rectangle\\
		9 & minorVrSpread & minor ellipse axis spread / $v_r$ spread\\		
		10 & majorVrSpread & major ellipse axis spread / $v_r$ spread\\
		11 & maxDistDev & mean target dist. to max. distance line\\	
		12 & rMean & mean target distance (range)\\
		13 & CBOmiddle & cumulative binary occupancy (mid. bin)\\
		14 & phiSpread & difference from max. to min. angle\\
		15 & rStd & standard deviation of target distance\\
		16 & phiStd & standard deviation of azimuth angle\\
		17 & covEV1 & top eigenvalue of cov. of $x$ and $y$\\
		18 & ampStd & standard deviation of target amplitudes\\
		19 & cohuPerimeter & perimeter of min. convex cluster hull\\
		20 & con95minor & minor axis length of 95\% conf. ellipse\\
		21 & compactness & mean target distance to cluster center\\
		22 & vrStd & standard deviation of radial velocity\\
		23 & cohuDensity & ratio of targets to convex hull area\\
		24 & xyLinearity & linear correlation between $x$ and $y$\\
		25 & fitCircleRadius & radius of the best fitting circle \\
		26 & rehuDensity & ratio of targets to bound. rect. area\\
		27 & cohuArea & area of convex cluster hull \\
		28 & circularity & isoperimetric quotient of the convex hull\\
		29 & rehuArea & area of min. bounding rectangle \\
		30 & vrCompMin & min. radial velocity (ego-motion comp.)\\
		31 & phiVrLinearity & linear correlation between $\phi$ and $v_r$\\
		32 & phiSpreadComp & phiSpread weighted by mean distance \\
		33 & rMax & maximum target distance (range)\\
		34 & nTargets & total amount of targets in cluster \\
		35 & rVrLinearity & linear correlation between $r$ and $v_r$\\
		36 & covEV2 & lower eigenvalue of cov. of $x$ and $y$\\
		\hline
	\end{tabular}
\end{table}

Second, 10-fold cross-validation was used to evaluate the classification performance of each data subset. The subsets start with the ten top ranked features and consecutively add the next best features, five at a time for a first preliminary search. After this coarse search a finer step size of just one feature was used to search for a final setting arround the best performing preliminary set. Results were averaged over all folds and the subset with the highest score was selected for each method. This process was repeated twice, once using normal multiclass models and the second time for the PWC-OVA method as this yielded the best results among all binarization techniques. These experiments show that the binarized classifier models are better equipped to make use of a larger number of features. A natural reason for this behavior would be the largely increased number of trees or LSTM cells which comprise a PWC or OVA model. Similar results could not be reproduced with an increased number of trees or cells and a multiclass classifier, though. The best subset for backward elimination and random forests consists of 22 features (F1: $\SI{88.40}{\%}$) in the multiclass and 31 for PWC-OVA (F1: $\SI{88.74}{\%}$). LSTM achieve their best backward elimination results using also 22 features (F1: $\SI{89.21}{\%}$) and 36 for PWC-OVA (F1: $\SI{90.24}{\%}$). Both results clearly demonstrate the advantage of the additional features when compared to the results in \cite{schumann17}, where 34 features lead to F1-scores of $\SI{87.1}{\%}$ (random forest) and $\SI{88.4}{\%}$ (LSTM) in the multiclass case.

\subsection{Imbalance Treatments}
\begin{figure}[h]
	\centering
	\includegraphics[width=1.\columnwidth]{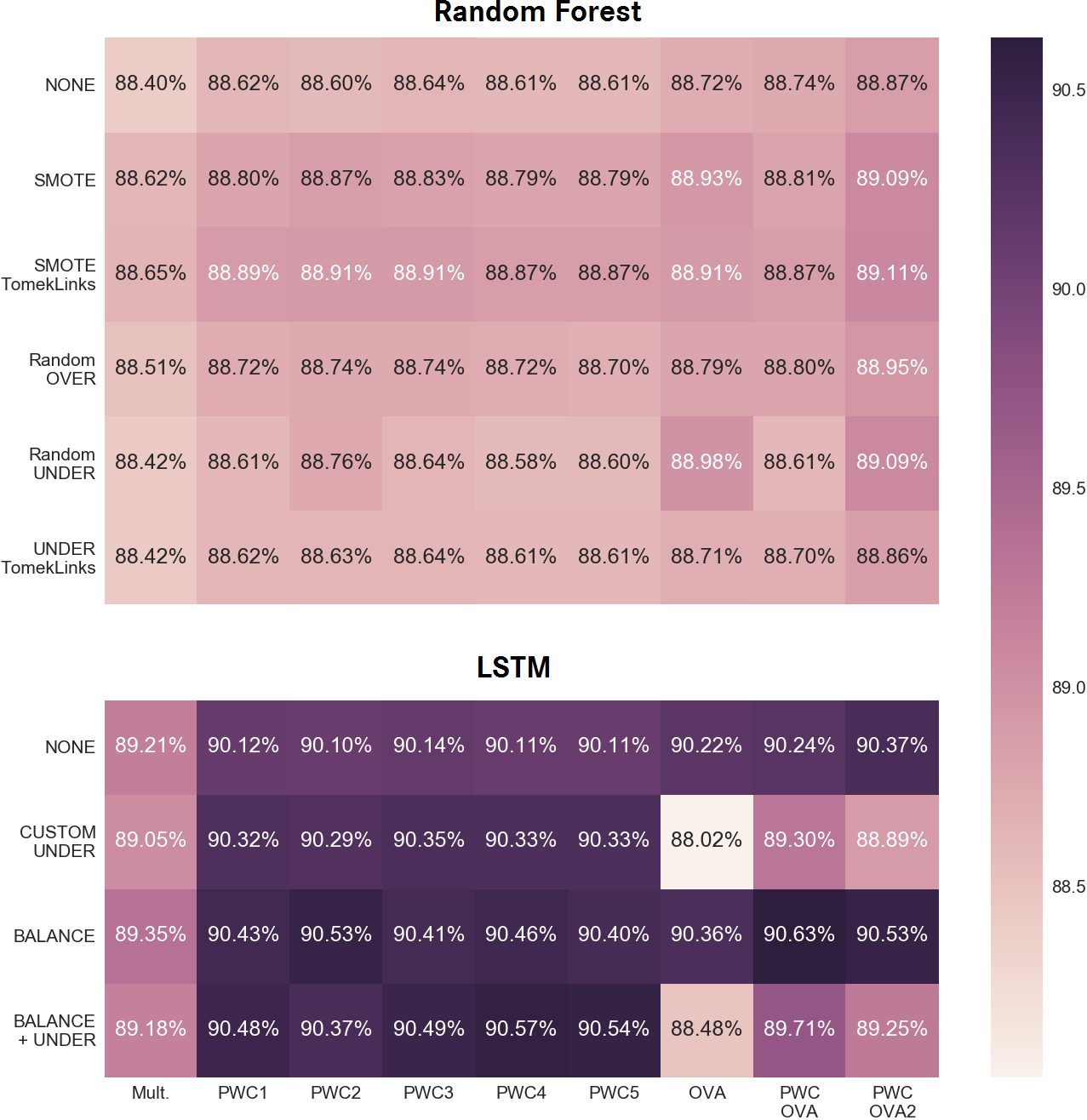}
	\caption{Validation scores for different re-sampling techniques evaluated with varying multiclass classification methods. Random Forest results are displayed on top, LSTM scores are given at the bottom.}	
	\label{fig:gridsearch}
\end{figure}

To counteract balance issues in the dataset, all binarization techniques described in Section \ref{seq:methods} were examined including the classical multiclass method. A grid search with random forests was used to test all combinations of multiclass classification methods and the mentioned re-sampling and weighting techniques. Results have been gathered using a 10x10-fold nested cross-validation. This was possible because many of the PWC techniques utilize the same binary classifiers, therefore all these models have to be trained only once per fold. Nested cross-validation is necessary in order to get unbiased results, i.e., if one part of the data is used to decide for the best among several classification techniques the same data should not be used as basis for a final performance score. Decision scores of methods with beneficial results during the random forest grid search can be found in Fig. \ref{fig:gridsearch}. Class weighting and oversampling with ADASYN are not displayed because these methods do not lead to improved classification scores.

Overall, the evaluation of the random forest grid search shows that techniques involving one-vs-all classifiers, i.e., OVA, PWC-OVA, and PWC-OVA2, have similar high scores. Among the different PWC-x weighting techniques with no correcting classifiers PWC-2 gives the best result (except for the combination with no re-sampling), closely followed by PWC-3 and the adapted version of PWC-1. Most importantly, it is clearly visible that the overall score improves for all binarization techniques when compared to the normal multiclass case. Among the sample balancing methods, SMOTE and a SMOTE variant using undersampling with Tomek links removal performed consistently well for all multiclass methods. Also, some benefits can be seen for certain combinations with undersampling (random and using Tomek links) and random oversampling. Best results are obtained using PWC-OVA2 with SMOTE and Tomek link removal.

As LSTM training requires substantially more time, only 10-fold cross-validation with a subsequent 80/20 split on every fold was used in order to get independent test and validation sets. For LSTMs only class weighting and random under- and oversampling were tested since more specialized techniques such as synthetic sample design cannot be easily applied to time sequences and are beyond the scope of this work. A standard implementation of over- or undersampling leads to drastic classification score deterioration, hence those techniques were not incorporated in the final grid search. Utilizing a custom undersampling ratio that only drops samples down to a minimum of five times the smallest class size, it is possible to improve the F1-scores by roughly 0.2 percentage points for all PWC-x methods and even further when combined with class weighting. Even with this custom ratio, undersampling still leads to worse results for any technique involving OVA classifiers. As shown in Fig. \ref{fig:gridsearch}, the best LSTM performance for all binarization methods can be gained by applying class weighting. Best overall results are obtained in combination with PWC-OVA. In general, the potential of multiclass binarization is much more distinct for LSTM classifiers which usually improve the F1-score by more than 0.8 percentage points except for the combinations of OVA and undersampling.

The final scores on the test sets of selected methods can be found in Tab. \ref{tab:res_scores}. The list contains the F1-scores for the optimal settings (displayed in bold font) separated for random forests and LSTMs. Each classifier is evaluated with is optimal feature set as identified in the previous subsection. For comparative purposes also the results obtained from plain multiclass classification and PWC-OVA2 are listed.

\begin{table}[h]
\renewcommand{\arraystretch}{1.3}
	\caption{Selection of final classification results on test set using optimum feature sets (top scores highlighted).}
	\label{tab:res_scores}
	\centering
	\begin{tabular}{lc}
		\hline
		Random Forest & $F1_\text{macro\_avg}$ \\
		\hline\hline
		Multiclass (No Re-Sampling or Weighting) & 0.8838  $\pm$ 0.001 \\
		PWC-OVA2 (No Re-Sampling or Weighting) & 0.8884  $\pm$ 0.001 \\
		PWC-OVA2 (SMOTE + Tomek Links Undersamp.) & \textbf{0.8912  $\pm$ 0.001}  \\
		\hline
		LSTM & $F1_\text{macro\_avg}$ \\
		\hline\hline
		Multiclass (No Re-Sampling or Weighting) & 0.8928 $\pm$ 0.002 \\
		PWC-OVA (Class Weighting) & \textbf{0.9059 $\pm$ 0.003} \\
		PWC-OVA2 (Class Weighting) & 0.9041 $\pm$ 0.003 \\
		\hline
	\end{tabular}
\end{table}

\begin{figure}[h]
	\centering
	\includegraphics[width=0.95\columnwidth]{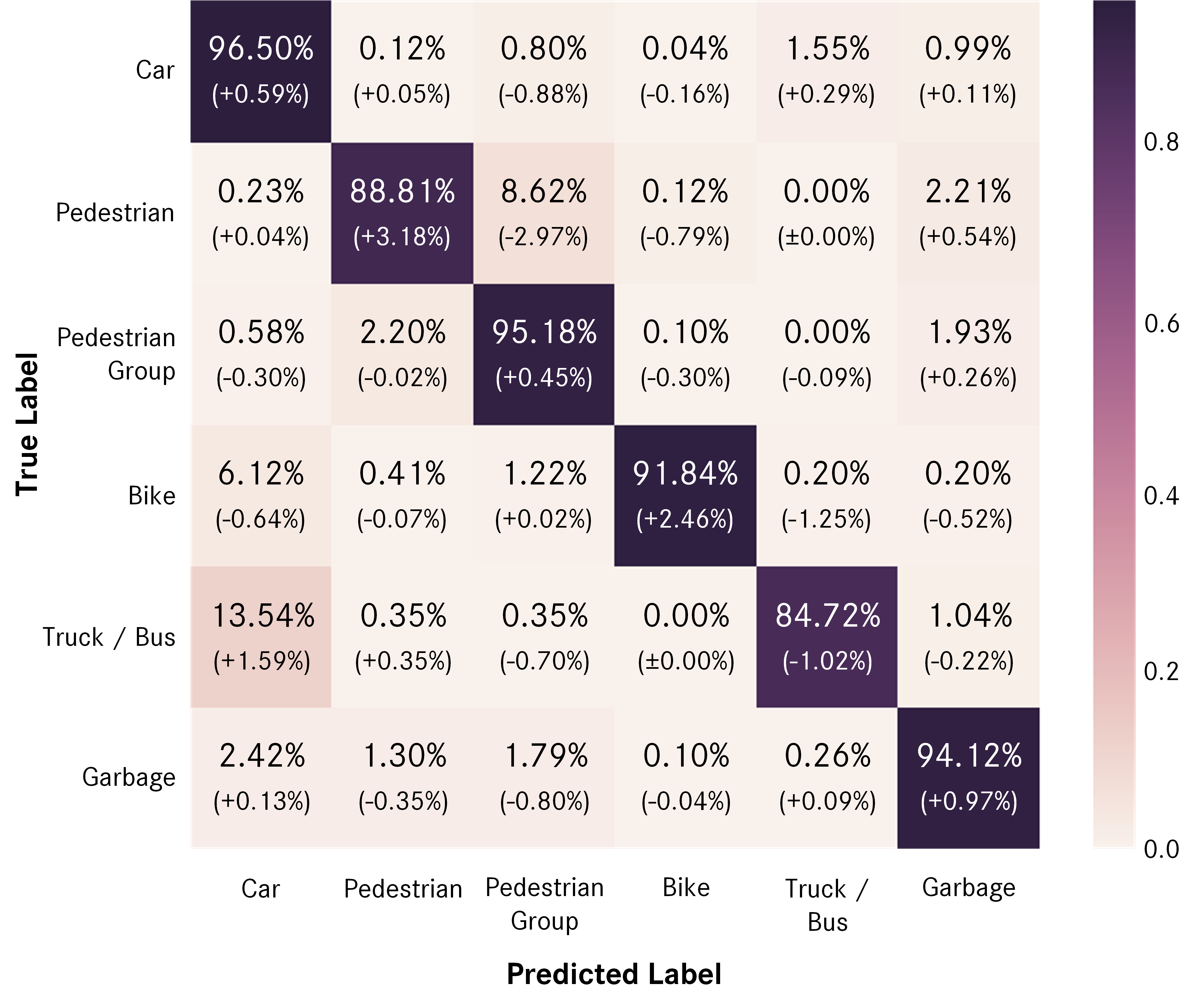}
	\caption{Confusion matrix for LSTM classifier using PWC-OVA with class weighting. Correct classifications are located along the main diagonal. The differences to normal multiclass LSTM classification is given in brackets.}	
	\label{fig:cm}
\end{figure}

The initial claim of this article was that using binarization techniques instead of the classical multiclass approach would help dealing with class imbalances, especially by improving the scores on minority classes. To this end, Fig. \ref{fig:cm} depicts the confusion matrix for the results obtained from the best configuration found in this work, i.e., the LSTM using the class-weighted PWC-OVA method. To make the results comparable, the figure also shows the differences towards the multiclass LSTM approach.

As expected, the biggest improvements with PWC-OVA are achieved on two of the minority classes, bike and pedestrian, as indicated by Fig. \ref{fig:cm}. But also for all other classes except the truck/bus class improvements are made along the main diagonal of the confusion matrix. The deterioration in truck recognition is probably observed because trucks and buses can in general look a lot like cars from a radar's perspective. This is caused by the fact, that often only the front or rear edges of a vehicle are captured by the sensors resulting in two smaller clusters. Without height information it is very hard to distinguish these objects. As a result, many of the binary classifiers that distinguish cars from any objects other than trucks do now give a strong indication towards the car class and the important car vs. truck/bus classifier may not have enough impact to steer the decision towards truck.
This knowledge can in fact be used for a two stage classification algorithm that uses PWC-OVA in a first stage. When deciding for a car in the first stage, the classifier then uses the coupled probability $p_{\texttt{truck/bus,car}}(\bm{x})$ to switch the final output to truck when $p_{\texttt{truck/bus,car}}(\bm{x}) > \texttt{thr}$, where $\texttt{thr}$ is a variable threshold level. Experiments with this setup show that the positive effects on the car class as depicted in Fig. \ref{fig:cm} can actually be exchanged for the loss in truck/bus recognition, depending on the adjustment of $\texttt{thr}$. For $\texttt{thr} = 0.75$ the overall test result is an F1-score of $\SI{90.57}{\%}$ which is slightly worse than for pure PWC-OVA, but in return a resulting truck accuracy of $\SI{85.95}{\%}$ upholds the statement, that all minority class scores can be improved.

\section{Conclusion} \label{seq:conc}
In this article, a motivation for the classification of traffic participants with automotive radar was given. The dataset used for the experiments was described and it was shown how the data can be structured and features can be extracted for a subsequent classification task. It was shown that geometric fits such as the minimum rectangular bounding box or micro-Doppler features such as range-Doppler spread are good additions to standard sensor output descriptors. The extracted feature sets were downsized, i.e., suitable feature subsets were evaluated and passed over to random forest and LSTM algorithms. As the data suffers from imbalance problems, classifier binarization was applied to enhance classification performance, especially for minority classes. This is further supported by class weighting and different over- and undersampling methods, which aim to restore data balance for the binary models at training time. With few exceptions on special combinations, final results showed a substantial improvement when using any of the binarization techniques.
A in-depth test for statistical significance is planned for future works as well as a performance evaluation on a publicly available dataset, which unfortunately is not existent for automotive radar data at the moment.
The random forests used in this article yielded their best results by applying the newly proposed PWC-OVA2 methods. Best overall results were obtained using a LSTM classifier with class weighting and ordinary PWC-OVA, hence pairwise coupling with correction classifiers obtained from an one-vs-all training. The achieved results are highly competitive.
Automotive radar classification of moving traffic participants is mostly focused on fewer object classes or achieves worse results on minority groups. The small amount of training samples still limits the classification potential when compared to classical image-based approaches. However, it is expected that sensor fusion can profit a lot from a radar system with the proposed processing. In future works it is planned to enhance current results using radar sensors with appropriate signal processing techniques that allow to increase the resolution in range, angle, Doppler, and time. With an increasing number of detected radar targets, it is expected to get even better classification results. This is mainly motivated by the observation that spatially small objects such as pedestrians are often only reflected as one or two targets at a time step, which makes it very hard for these cases to calculate accurate features for those objects.

\section{Acknowledgment}
The research leading to these results has received funding from the European Union under the H2020 ECSEL Programme as part of the DENSE project, contract number 692449.

\bibliographystyle{IEEEtran}
\bibliography{IEEEabrv,mybibfile}

\begin{thebibliography}{10}
\providecommand{\url}[1]{#1}
\csname url@rmstyle\endcsname
\providecommand{\newblock}{\relax}
\providecommand{\bibinfo}[2]{#2}
\providecommand\BIBentrySTDinterwordspacing{\spaceskip=0pt\relax}
\providecommand\BIBentryALTinterwordstretchfactor{4}
\providecommand\BIBentryALTinterwordspacing{\spaceskip=\fontdimen2\font plus
\BIBentryALTinterwordstretchfactor\fontdimen3\font minus
  \fontdimen4\font\relax}
\providecommand\BIBforeignlanguage[2]{{%
\expandafter\ifx\csname l@#1\endcsname\relax
\typeout{** WARNING: IEEEtran.bst: No hyphenation pattern has been}%
\typeout{** loaded for the language `#1'. Using the pattern for}%
\typeout{** the default language instead.}%
\else
\language=\csname l@#1\endcsname
\fi
#2}}

\bibitem{schubert15}
E.~Schubert, F.~Meinl, M.~Kunert, and W.~Menzel, ``{High resolution automotive
  radar measurements of vulnerable road users - pedestrians {\&} cyclists},''
  in \emph{2015 IEEE MTT-S International Conference on Microwaves for
  Intelligent Mobility (ICMIM)}, no. April.\hskip 1em plus 0.5em minus
  0.4em\relax IEEE, apr 2015, pp. 1--4.

\bibitem{molchanov14}
P.~Molchanov, ``{Radar Target Classification by Micro-Doppler Contributions},''
  PhD Thesis, Tampere University of Technology, 2014.

\bibitem{hasch12}
J.~Hasch, E.~Topak, R.~Schnabel, T.~Zwick, R.~Weigel, and C.~Waldschmidt,
  ``Millimeter-wave technology for automotive radar sensors in the 77 ghz
  frequency band,'' \emph{IEEE Transactions on Microwave Theory and
  Techniques}, vol.~60, no.~3, pp. 845--860, March 2012.

\bibitem{lecun98}
Y.~LeCun, L.~Bottou, Y.~Bengio, and P.~Haffner, ``{Gradient-based learning
  applied to document recognition},'' \emph{Proceedings of the IEEE}, vol.~86,
  no.~11, pp. 2278--2323, 1998.

\bibitem{alexnet12}
A.~Krizhevsky, I.~Sutskever, and G.~E. Hinton, ``{ImageNet Classification with
  Deep Convolutional Neural Networks},'' in \emph{2012 25th International
  Conference on Neural Information Processing Systems (NIPS)}, vol.~1, 2012,
  pp. 1097--1105.

\bibitem{vgg14}
K.~Simonyan and A.~Zisserman, ``{Very Deep Convolutional Networks for
  Large-Scale Image Recognition},'' \emph{CoRR}, vol. abs/1409.1, pp. 1--14,
  2014.

\bibitem{lomb15}
J.~Lombacher, M.~Hahn, J.~Dickmann, and C.~W{\"{o}}hler, ``{Detection of
  arbitrarily rotated parked cars based on radar sensors},'' in \emph{2015 16th
  International Radar Symposium (IRS)}.\hskip 1em plus 0.5em minus 0.4em\relax
  IEEE, jun 2015, pp. 180--185.

\bibitem{lomb17}
------, ``{Object Classification in Radar Using Ensemble Methods},'' in
  \emph{2017 IEEE MTT-S International Conference on Microwaves for Intelligent
  Mobility (ICMIM)}, 2017, pp. 87--90.

\bibitem{bartsch12}
A.~Bartsch, F.~Fitzek, and R.~H. Rasshofer, ``{Pedestrian recognition using
  automotive radar sensors},'' \emph{Advances in Radio Science}, vol.~10, pp.
  45--55, sep 2012.

\bibitem{heuel11}
S.~Heuel and H.~Rohling, ``{Two-stage pedestrian classification in automotive
  radar systems},'' in \emph{2011 12th International Radar Symposium (IRS)},
  sep 2011, pp. 477--484.

\bibitem{heuel13}
------, ``{Pedestrian Recognition Based on 24 GHz Radar Sensors},'' in
  \emph{Ultra-Wideband Radio Technologies for Communications, Localization and
  Sensor Applications}, R.~S. Thom{\"{a}}, Ed.\hskip 1em plus 0.5em minus
  0.4em\relax InTech, 2013, ch.~10, pp. 241--256.

\bibitem{schumann17}
O.~Schumann, M.~Hahn, J.~Dickmann, and C.~W{\"{o}}hler, ``{Comparison of Random
  Forest and Long Short-Term Memory Network Performances in Classification
  Tasks Using Radar},'' in \emph{2017 Symposium Sensor Data Fusion 2017
  (SSDF)}, Bonn, Germany, 2017.

\bibitem{richards2005}
M.~A. Richards, \emph{Fundamentals of Radar Signal Processing}, ser.
  Professional Engineering.\hskip 1em plus 0.5em minus 0.4em\relax McGraw-Hill,
  2005.

\bibitem{ester96}
M.~Ester, H.-P. Kriegel, J.~Sander, and X.~Xu, ``{A Density-based Algorithm for
  Discovering Clusters in Large Spatial Databases with Noise},'' \emph{1996 2nd
  International Conference on Knowledge Discovery and Data Mining (KDD)}, pp.
  226--231, aug 1996.

\bibitem{breiman01}
L.~Breiman, ``{Random Forests},'' \emph{Machine Learning}, vol.~45, no.~1, pp.
  5--32, oct 2001.

\bibitem{hochreiter97}
S.~Hochreiter and J.~Schmidhuber, ``{Long Short-Term Memory},'' \emph{Neural
  Computation}, vol.~9, no.~8, pp. 1735--1780, Nov. 1997.

\bibitem{zhang16}
X.~Zhang, ``{Dynamic Object Classification Based on High Resolution Doppler
  Radar and Laser Data},'' Master's Thesis, Universit{\"{a}}t Stuttgart, 2016.

\bibitem{Kohavi1997}
R.~Kohavi and G.~H. John, ``{Wrappers for Feature Subset Selection},''
  \emph{Artificial Intelligence}, vol.~97, no. 1-2, pp. 273--324, dec 1997.

\bibitem{fernandez13}
A.~Fern{\'{a}}ndez, V.~L{\'{o}}pez, M.~Galar, M.~J. {Del Jesus}, and
  F.~Herrera, ``{Analysing the classification of imbalanced data-sets with
  multiple classes: Binarization techniques and ad-hoc approaches},''
  \emph{Knowledge-Based Systems}, vol.~42, pp. 97--110, 2013.

\bibitem{chmielnicki2016}
W.~Chmielnicki and K.~Stapor, ``{Combining One-Versus-One and One-Versus-All
  Strategies to Improve Multiclass SVM Classifier},'' in \emph{2015 9th
  International Conference on Computer Recognition Systems (CORES)}, R.~Burduk,
  K.~Jackowski, M.~Kurzy{\'{n}}ski, M.~Wo{\'{z}}niak, and
  A.~{\.{Z}}o{\l}nierek, Eds., vol. 403.\hskip 1em plus 0.5em minus 0.4em\relax
  Springer International Publishing, 2016, pp. 37--45.

\bibitem{friedman96}
J.~H. Friedman, ``{Another Approach to Polychotomous Classification},''
  Department of Statistics and Stanford Linear Accelerator Center, Stanford
  University, Stanford, California, USA, Tech. Rep., 1996.

\bibitem{moreira98}
M.~Moreira and E.~Mayoraz, ``{Improved pairwise coupling classification with
  correcting classifiers},'' in \emph{1998 10th European Conference on Machine
  Learning (ECML-98)}, C.~N{\'{e}}dellec and C.~Rouveirol, Eds.\hskip 1em plus
  0.5em minus 0.4em\relax Chemnitz, Germany: Springer Berlin Heidelberg, apr
  1998, pp. 160--171.

\bibitem{schlegel2017}
B.~Schlegel and B.~Sick, ``Dealing with class imbalance the scalable way:
  Evaluation of various techniques based on classification grade and
  computational complexity,'' in \emph{2017 IEEE International Conference on
  Data Mining Workshops (ICDMW)}, Nov 2017, pp. 69--78.

\end{thebibliography}

\end{document}